\documentclass[conference]{IEEEtran}
\IEEEoverridecommandlockouts
\usepackage{cite}
\usepackage{amsmath,amssymb,amsfonts}
\usepackage{hyperref}
\usepackage{algorithmic}
\usepackage{graphicx}
\usepackage{textcomp}
\newcounter{wikiseedfn}
\usepackage{subcaption}
\usepackage{xcolor}
\def\BibTeX{{\rm B\kern-.05em{\sc i\kern-.025em b}\kern-.08em
    T\kern-.1667em\lower.7ex\hbox{E}\kern-.125emX}}
\begin{document}

\title{Fine-Tuning vs. RAG for Multi-Hop Question Answering with Novel Knowledge}

\author{
\IEEEauthorblockN{
Zhuoyi Yang,
Yurun Song,
Iftekhar Ahmed,
Ian Harris
}
\IEEEauthorblockA{
University of California, Irvine\\
Irvine, California, USA\\
\{zhuoyy1, yuruns, iftekha, iharris\}@uci.edu
}
}

\maketitle

\begin{abstract}
Multi-hop question answering is widely used to evaluate the reasoning capabilities of large language models (LLMs), as it requires integrating multiple pieces of supporting knowledge to arrive at a correct answer. While prior work has explored different mechanisms for providing knowledge to LLMs, such as finetuning and retrieval-augmented generation (RAG), their relative effectiveness for multi-hop question answering remains insufficiently understood, particularly when the required knowledge is temporally novel.

In this paper, we systematically compare parametric and non-parametric knowledge injection methods for open-domain multi-hop question answering. We evaluate unsupervised fine-tuning (continual pretraining), supervised fine-tuning, and retrieval-augmented generation across three 7B-parameter open-source LLMs. Experiments are conducted on two benchmarks: QASC, a standard multi-hop science question answering dataset, and a newly constructed dataset of over 10,000 multi-hop questions derived from Wikipedia events in 2024, designed to test knowledge beyond the models’ pretraining cutoff.

Our results show that unsupervised fine-tuning provides only limited gains over base models, suggesting that continual pretraining alone is insufficient for improving multi-hop reasoning accuracy. In contrast, retrieval-augmented generation yields substantial and consistent improvements, particularly when answering questions that rely on temporally novel information. Supervised fine-tuning achieves the highest overall accuracy across models and datasets. These findings highlight fundamental differences in how knowledge injection mechanisms support multi-hop question answering and underscore the importance of retrieval-based methods when external or compositional knowledge is required.
\end{abstract}


\section{Introduction}
Large language models (LLMs) have demonstrated strong performance on a wide range of question answering tasks, including those that require combining information from multiple sources \cite{hotpotqa}. Among these tasks, multi-hop question answering has emerged as a key benchmark for evaluating a model’s ability to integrate multiple pieces of evidence and perform compositional reasoning \cite{2wikimultihopqa}. Rather than relying on a single fact, multi-hop questions require models to connect several supporting facts—often drawn from different documents or distant parts of a text—to arrive at the correct answer\cite{musique}.

One key factor that influences multi-hop question answering performance is the mechanism used to inject knowledge into the model. Broadly, existing approaches fall into two categories. Parametric methods incorporate knowledge directly into model parameters through wight updates, either using unlabeled text (unsupervised fine-tuning or continual pretraining)\cite{dontstopPretraining} or labeled question–answer pairs (supervised fine-tuning) \cite{unifiedqa}. In contrast, non-parametric methods such as retrieval-augmented generation (RAG) provide external evidence at inference time by retrieving relevant documents from a knowledge corpus and conditioning the model’s predictions on the retrieved text\cite{rag}.

Prior work has compared fine-tuning and RAG primarily in the context of factual recall and single-hop question answering, often finding retrieval-based methods to be competitive or superior when external knowledge is required \cite{singleHop_finetuningVSrag}. However, it remains unclear whether these conclusions extend to multi-hop question answering. Answering multi-hop questions requires processing longer contexts, identifying multiple relevant facts, and composing them coherently. In such settings, retrievers may fail to retrieve all necessary evidence, while frozen generators may struggle to effectively integrate retrieved information through in-context learning alone \cite{2wikimultihopqa,hotpotqa,musique}. 

In this paper, we aim to answer the following research question (RQ): Does the method of knowledge injection impact the effectiveness of open-domain multi-hop question answering? To address this question, we conducted a systematic comparison of finetuning and RAG under controlled experimental settings. We consider unsupervised finetuning, supervised finetuning, and RAG as three distinct knowledge injection mechanisms that differ in when and how supporting knowledge is made available to the model.

We evaluate these methods on two multi-hop benchmarks that represent different knowledge conditions. The first is QASC dataset \cite{qascDataset}, a widely used multi-hop science question answering dataset. The second is 2024 Events dataset, a newly constructed dataset of over 10,000 multi-hop questions derived from Wikipedia events in 2024 from Wikipedia events in 2024%
\footnote{\url{https://en.wikipedia.org/wiki/Category:2024_in_the_United_States_by_month}}
\setcounter{wikiseedfn}{\value{footnote}}, designed to assess performance under temporally novel knowledge that is not memorized during pretraining. Using three open-source 7B-parameter LLMs, we compare answer accuracy across all settings using a unified multiple-choice evaluation framework.

Our experimental results show consistent trends across both datasets. Unsupervised fine-tuning provides only marginal improvements over base models, suggesting that continual pretraining alone is insufficient for improving multi-hop reasoning accuracy. Retrieval-augmented generation substantially improves performance, more than doubling accuracy on the 2024 Events dataset. Supervised fine-tuning achieves the highest overall accuracy, highlighting the strong effect of task-specific supervision. \textbf{We made the following contributions:}

\begin{itemize}
    \item We analyze the relative effectiveness of retrieval-based versus parameter-based knowledge injection methods for reasoning, highlighting their impact on accuracy and generalization across datasets.

  \item We study multi-hop question answering, a more challenging setting than single-hop question answering, where answering requires integrating multiple pieces of information and reasoning over a larger body of knowledge.

   \item We study how LLMs perform when answering questions that rely on novel knowledge introduced after their pre-training cutoff, for which the models have prior exposure.
\end{itemize}

\section{Related Work}

\subsection{Multi-hop question Answering as a Reasoning Benchmark}

Multi-hop question answering has been widely adopted as a benchmark for evaluating a model’s ability to combine information from multiple sources. Datasets such as QASC, HotpotQA, 2WikiMultiHopQA, and MuSiQue are specifically designed to require evidence aggregation across multiple facts, sentences, or documents. Because these tasks cannot typically be solved using a single retrieved fact, performance on multi-hop benchmarks is often interpreted as a proxy for a model’s reasoning capability \cite{hotpotqa,musique,qascDataset,2wikimultihopqa}.

Most prior work evaluates multi-hop question answering systems using answer accuracy as the primary metric, without explicitly supervising or inspecting intermediate reasoning steps \cite{hotpotqa}. As a result, improvements in accuracy are often attributed to better reasoning, although they may also reflect gains in retrieval quality, memorization, or task-specific heuristics \cite{stochastic}. This limitation has motivated recent studies to examine how architectural choices, prompting strategies, and training objectives influence multi-hop performance, even when explicit reasoning traces are not available \cite{cot}.

Our work follows this evaluation paradigm by using accuracy as the primary metric \cite{singleHop_finetuningVSrag}, but focuses specifically on how different knowledge injection mechanisms affect multi-hop question answering outcomes. Rather than proposing a new reasoning architecture or dataset, we study how models behave under different ways of accessing supporting knowledge, holding model scale and evaluation format constant. 

\subsection{Knowledge Injection in Large Language Models}
Large language models acquire substantial factual knowledge during pretraining \cite{Language_Models_as_Knowledge_Bases}, but this knowledge is inherently static and bounded by the training corpus \cite{gpt3}. To address these limitations, prior work has explored various mechanisms for injecting additional knowledge into LLMs, including continual pretraining \cite{dontstopPretraining}, supervised fine-tuning \cite{unifiedqa}, and retrieval-based methods \cite{flare,Self-rag,ChainOFrag}. These approaches differ fundamentally in whether knowledge is encoded parametrically in model weights or provided dynamically at inference time.

Unsupervised fine-tuning, also referred to as continual pretraining, incorporates new information by further training a model on unlabeled text using a language modeling objective. This approach is appealing due to its scalability and lack of annotation requirements. However, prior studies have shown that continual pretraining struggles to internalize sparse or temporally novel facts, and its benefits for downstream reasoning tasks are often limited \cite{rag,dontstopPretraining}. In the context of multi-hop question answering, this method requires the model to implicitly encode all relevant supporting knowledge within its parameters prior to inference.

Supervised fine-tuning injects knowledge through labeled examples, typically in the form of question–answer pairs or instruction-following data. This approach has been shown to substantially improve downstream task performance and alignment with evaluation formats \cite{gpt3, Training_Language_Models_to_Follow_Instructions_with_Human_Feedback}. However, supervised fine-tuning may conflate knowledge acquisition with task-specific pattern learning, making it difficult to disentangle improvements due to reasoning ability from those due to answer format adaptation or shortcut learning \cite{shortcut_learning}. While widely used, supervised fine-tuning does not directly address how models access or combine external knowledge at inference time.

Retrieval-augmented generation (RAG) represents a non-parametric alternative, in which relevant documents are retrieved from an external corpus and provided to the model as additional context during inference \cite{rag}. By decoupling knowledge storage from model parameters, RAG enables access to large and dynamically updated knowledge sources and has been shown to be effective for factual recall and knowledge-intensive tasks. Several studies have demonstrated strong RAG performance on single-hop or weakly compositional question answering tasks. However, less is known about how retrieval quality and context integration affect performance on multi-hop questions that require composing multiple pieces of evidence \cite{singleHop_finetuningVSrag, flare,Self-rag,ChainOFrag}.

Our work builds on this literature by directly comparing unsupervised fine-tuning, supervised fine-tuning, and retrieval-augmented generation under a unified evaluation framework. Unlike prior studies that focus primarily on single-hop factual recall or open-ended generation, we examine how these knowledge injection mechanisms influence multiple-choice multi-hop question answering, including settings involving temporally novel knowledge.






\begin{figure*}[t]
  \centering
    \includegraphics[width=\textwidth]{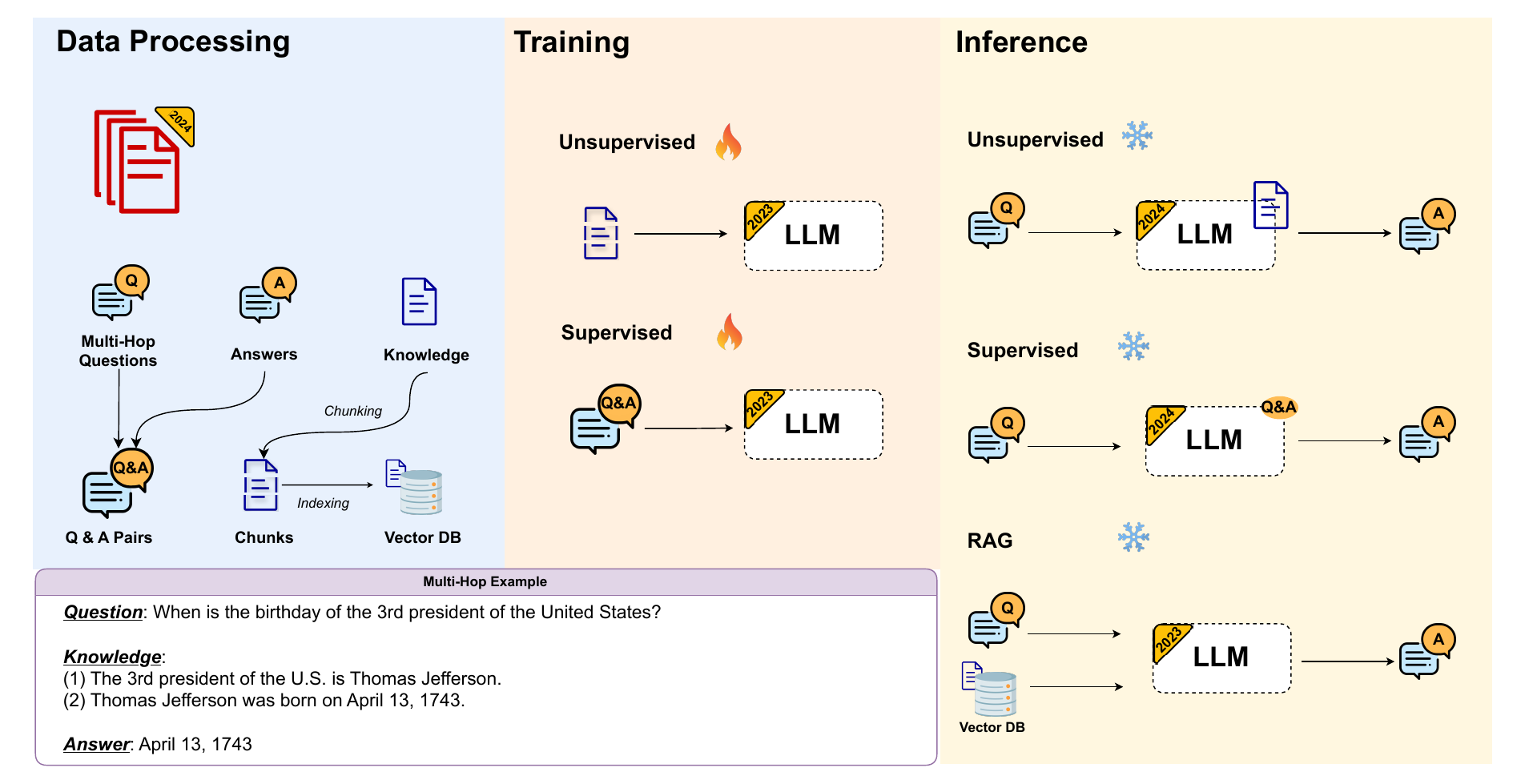}
  \caption{Comparative Framework for Multi-Hop Question Answering across Knowledge Injection Methods}
  \label{overview}
\end{figure*}

\section{Overview}
Figure \ref{overview} provides an overview of our experimental framework for studying multi-hop question answering under different knowledge injection mechanisms. To enable a controlled comparison, we construct both the \textbf{knowledge base} and \textbf{benchmark datasets} consisting of multi-hop questions with corresponding answers. We consider three representative knowledge injection strategies: \textbf{unsupervised fine-tuning, supervised fine-tuning, and retrieval-augmented generation (RAG)}. 

For unsupervised finetuning, knowledge is incorporated into the model via unlabeled text. For supervised finetuning, knowledge is embedded into the model through question-answer pairs. In contrast, RAG injects external knowledge at inference time by retrieving relevant evidence from an external corpus. Using a unified multiple-choice evaluation setup, we assess the performance of all three mechanisms in terms of answer accuracy on the same benchmarks.

\section{Knowledge Base Construction}

\subsection{Benchmark Selection and Rationale}
We selected QASC Dataset and created our own 2024 Events Dataset from Wikipedia content.

\textbf{QASC Dataset} \cite{qascDataset} \hspace{0.5cm}
QASC (Question Answering via Sentence Composition) is a widely-used multi-hop reasoning dataset. Each question has eight answer options, among which only one is correct. We chose this dataset because it includes questions from different fields of science, which enables a more thorough evaluation of the model’s reasoning capability. However, we acknowledge that answers to questions in this dataset may have been partially observed during pre-training.

\textbf{2024 Events Dataset} \hspace{0.5cm}
We constructed a dataset of over 10,000 multi-hop multiple-choice questions from events that happened in 2024 based on Wikipedia content\footnotemark[\value{wikiseedfn}]. Each question has four answer options with only one being correct. We describe how this dataset is contructed in details in Section \ref{data collection and knowledge corpora}. This design enables the evaluation of multi-hop question answering under temporally novel knowledge conditions. Since these articles postdate the training cutoff of our evaluated models, they are not memorized by the models during pre-training.

\subsection{Data Collection and Knowledge Corpora}
\label{data collection and knowledge corpora}
\textbf{QASC Dataset} \hspace{0.5cm} Following prior work \cite{arc}, we categorized questions in this dataset into seven scientific domains: Physics, Chemistry, Biology, Earth Science, Astronomy, Environmental Science, and General Science. We use GPT-4 \cite{openai2024gpt4technicalreport} to label each question accordingly.

We used the external knowledge corpus provided with the QASC dataset \cite{qascDataset}, which consists of approximately 17 million science-related sentences across multiple domains. This fixed corpus was built prior to evaluation and serves as both unsupervised finetuning training corpus and the retrieval source.

\textbf{2024 Events Dataset} \hspace{0.5cm} 
For the 2024 events dataset, we constructed the knowledge corpus from scratch using Wikipedia. We used the Wikipedia category page “2024 in the United States by month\footnotemark[\value{wikiseedfn}]” as a seed and follow links to individual monthly pages. From each month, we extracted descriptions of real-world events that occurred during that period. All the events together formed our knowledge base.

To create multi-hop questions, we segmented each event description into chunks of 200 tokens to ensure manageable context lengths. For each chunk, we then prompted GPT-4 and DeepSeek-R1\cite{deepseek_r1} to independently generate two multi-hop multiple-choice questions, resulting in a diverse set of questions grounded in temporally novel information.

\begin{table*}[!t]
\centering
\begin{tabular}{llcccc}

\hline
Task                  & Model                                                                               & \multicolumn{1}{l}{Base model}                                & \multicolumn{1}{l}{Base model + RAG}                          & \multicolumn{1}{l}{Unsupervised Fine-tuned}                   & \multicolumn{1}{l}{Supervised Finetuned}                      \\ \hline
Physics               & \begin{tabular}[c]{@{}l@{}}Mistral-7B\\ Llama2-7B\\ Llama2-7B-Instruct\end{tabular} & \begin{tabular}[c]{@{}c@{}}0.354\\ 0.383\\ 0.457\end{tabular} & \begin{tabular}[c]{@{}c@{}}0.634\\ 0.659\\ 0.757\end{tabular} & \begin{tabular}[c]{@{}c@{}}0.362\\ 0.372\\ 0.449\end{tabular} & \begin{tabular}[c]{@{}c@{}}0.806\\ 0.812\\ 0.864\end{tabular} \\ \hline
Chemistry             & \begin{tabular}[c]{@{}l@{}}Mistral-7B\\ Llama2-7B\\ Llama2-7B-Instruct\end{tabular} & \begin{tabular}[c]{@{}c@{}}0.345\\ 0.351\\ 0.392\end{tabular} & \begin{tabular}[c]{@{}c@{}}0.623\\ 0.642\\ 0.735\end{tabular} & \begin{tabular}[c]{@{}c@{}}0.356\\ 0.343\\ 0.413\end{tabular} & \begin{tabular}[c]{@{}c@{}}0.812\\ 0.823\\ 0.874\end{tabular} \\ \hline
Biology               & \begin{tabular}[c]{@{}l@{}}Mistral-7B\\ Llama2-7B\\ Llama2-7B-Instruct\end{tabular} & \begin{tabular}[c]{@{}c@{}}0.353\\ 0.359\\ 0.401\end{tabular} & \begin{tabular}[c]{@{}c@{}}0.602\\ 0.614\\ 0.717\end{tabular} & \begin{tabular}[c]{@{}c@{}}0.367\\ 0.362\\ 0.409\end{tabular} & \begin{tabular}[c]{@{}c@{}}0.801\\ 0.818\\ 0.862\end{tabular} \\ \hline
Earth Science         & \begin{tabular}[c]{@{}l@{}}Mistral-7B\\ Llama2-7B\\ Llama2-7B-Instruct\end{tabular} & \begin{tabular}[c]{@{}c@{}}0.353\\ 0.367\\ 0.405\end{tabular} & \begin{tabular}[c]{@{}c@{}}0.604\\ 0.615\\ 0.748\end{tabular} & \begin{tabular}[c]{@{}c@{}}0.359\\ 0.379\\ 0.413\end{tabular} & \begin{tabular}[c]{@{}c@{}}0.812\\ 0.826\\ 0.861\end{tabular} \\ \hline
Astronomy             & \begin{tabular}[c]{@{}l@{}}Mistral-7B\\ Llama2-7B\\ Llama2-7B-Instruct\end{tabular} & \begin{tabular}[c]{@{}c@{}}0.362\\ 0.381\\ 0.410\end{tabular} & \begin{tabular}[c]{@{}c@{}}0.598\\ 0.634\\ 0.739\end{tabular} & \begin{tabular}[c]{@{}c@{}}0.375\\ 0.394\\ 0.427\end{tabular} & \begin{tabular}[c]{@{}c@{}}0.793\\ 0.791\\ 0.854\end{tabular} \\ \hline
Environmental Science & \begin{tabular}[c]{@{}l@{}}Mistral-7B\\ Llama2-7B\\ Llama2-7B-Instruct\end{tabular} & \begin{tabular}[c]{@{}c@{}}0.341\\ 0.351\\ 0.403\end{tabular} & \begin{tabular}[c]{@{}c@{}}0.595\\ 0.613\\ 0.735\end{tabular} & \begin{tabular}[c]{@{}c@{}}0.356\\ 0.369\\ 0.412\end{tabular} & \begin{tabular}[c]{@{}c@{}}0.803\\ 0.813\\ 0.868\end{tabular} \\ \hline
General Science       & \begin{tabular}[c]{@{}l@{}}Mistral-7B\\ Llama2-7B\\ Llama2-7B-Instruct\end{tabular} & \begin{tabular}[c]{@{}c@{}}0.352\\ 0.351\\ 0.400\end{tabular} & \begin{tabular}[c]{@{}c@{}}0.641\\ 0.654\\ 0.759\end{tabular} & \begin{tabular}[c]{@{}c@{}}0.365\\ 0.372\\ 0.414\end{tabular} & \begin{tabular}[c]{@{}c@{}}0.783\\ 0.815\\ 0.864\end{tabular} \\ \hline

\end{tabular}
\vspace{2mm}
\caption{Results for the QASC dataset in terms of accuracy}
\label{qasc results}
\end{table*}

\begin{table*}[!t]
\centering

\begin{tabular}{lcccc}
\hline
                  & \multicolumn{1}{l}{Base Model} & \multicolumn{1}{l}{Base Model + RAG} & \multicolumn{1}{l}{Unsupervised Finetuning} & \multicolumn{1}{l}{Supervised Finetuning} \\ \hline
Mistral-7B        & 0.276                          & 0.654                                & 0.320                                       & 0.813                                     \\ 
Llama2-7B         & 0.278                          & 0.672                                & 0.329                                       & 0.832                                     \\ 
Llama-7B-Instruct & 0.326                          & 0.753                                & 0.392                                       & 0.884                                     \\ \hline
\end{tabular}
\vspace{2mm}
\caption{Results for the 2024 Events dataset in terms of accuracy}
\label{new events results}
\end{table*}

\section{Experimental Setup}
\subsection{Model Selection}
We evaluate three open-source large language models with approximately 7 billion parameters: Mistral-7B, LLaMA-7B, and LLaMA-7B-Instruct \cite{mistral,llama,llama2_instruction}. Mistral-7B and LLaMA-7B are base pretrained models that have not undergone instruction tuning, and therefore provide a view of multi-hop question answering performance without explicit reasoning- or instruction-oriented alignment. In contrast, LLaMA-7B-Instruct has been instruction-tuned to better follow natural language prompts and perform reasoning-style tasks, and serves as a stronger upper bound on achievable performance. Using models with comparable parameter counts allows us to control for model scale while examining the impact of different knowledge access mechanisms on multi-hop question answering accuracy. All three models have pre-training data cutoff point before 2024.

\subsection{Implementation Details}
\subsubsection{Parametric Injection}
\textbf{Common Setup} \hspace{0.5cm}
All experiments were run on two NVIDIA A6000 GPUs. Unless otherwise stated, we initialized from the base model and apply LoRA to the query and value projections of each self-attention layer $(r=16,  \alpha=32, dropout=0.1) $, freezing all base parameters. We optimized with AdamW $ (lr=1e-5) $ using a linear decay schedule with 10\% warmup, trained for 20 epochs with bfloat16 mixed precision, batch size 16, and truncated sequences to 300 tokens. We used a 90\% training set and 10 \% evaluation set.

\textbf{Unsupervised Finetuning} \hspace{0.5cm}
We performed causal language modeling on raw text. For QASC dataset, we used the released external knowledge corpus (17M science sentences) and trained the model at the sentence level. For the 2024 Events setting, we scraped Wikipedia event articles as described in Section \ref{data collection and knowledge corpora} and segmented them into 1,000-token chunks with 200-token overlap. The model was trained using the standard autoregressive language modeling objective. 
Given a token sequence $(x_1, \ldots, x_T)$, we minimize:
\[
\mathcal{L}_{\text{CLM}} = - \sum_{t=1}^{T} \log p(x_t \mid x_{<t}).
\]

\textbf{Supervised Finetuning} \hspace{0.5cm}
We finetuned the model on multiple-choice question answering by concatenating the question with each candidate answer option and predicting the correct option using a classification head. We considered $C \in \{4, 8\}$ answer options (QASC: $C=8$; 2024 Events: $C=4$). For the 2024 Events dataset, we used DeepSeek-generated questions for training and GPT-generated questions for evaluation. The training objective minimizes the cross-entropy loss over the $C$ answer classes:
\[
\mathcal{L}_{\mathrm{CE}}
= - \log \frac{\exp(z_y)}{\sum_{c=1}^{C} \exp(z_c)},
\]
where $z \in \mathbb{R}^{C}$ denotes the output logits and $y \in \{1,\dots,C\}$ is the index of the correct answer option.

\subsubsection{RAG Pipeline}
\textbf{Knowledge Corpus and Indexing}
We constructed a knowledge corpus from a 2024 Wikipedia text dump as described in Section \ref{data collection and knowledge corpora}. Articles are segmented into overlapping chunks using a recursive character-based splitter \cite{Langchain} with a chunk size of 1000 tokens and an overlap of 300 tokens. Each chunk is embedded using the BGE-large-en dense embedding model\cite{BGE} and indexed with FAISS using inner-product similarity over L2-normalized vectors \cite{faiss}.

\textbf{Retrieval and Reranking} \hspace{0.5cm}
Given a question (optionally augmented with its title), we retrieve the top 20 candidate chunks using dense retriever BGE. These candidates are then reranked using a cross-encoder reranker (cross-encoder/ms-marco-MiniLM-L6-v2\cite{reranker}), and the top 4 chunks are selected as contextual evidence.

\textbf{Prompt Construction} \hspace{0.5cm}
The retrieved chunks are concatenated and prepended to the question and its answer options to form a single prompt. The model is instructed to answer using only the provided context and to output a single answer letter. To improve alignment with the desired behavior, we include three in-context examples in the prompt.

\textbf{Answer Scoring and Selection}\hspace{0.5cm}
Rather than generating free-form text, we adopt an MMLU-style scoring strategy \cite{mmlu}: Given the prompt, we compute the log-probability of generating each answer option token (A,B,C,D) at the next position. The option with the highest log-probability is selected as the final prediction. This approach ensures direct comparability with classification-based evaluation while avoiding ambiguity from open-ended generation.

\subsection{Evaluation Method}
All models were evaluated under a unified multiple-choice classification framework, where the goal was to select the only correct answer from option candidates. We used accuracy as the primary evaluation metric. Accuracy is defined as the fraction of questions for which the predicted answer label matches the ground-truth label:
\[
\text{Accuracy} = \frac{1}{N} \sum_{i=1}^{N} \mathbf{1}[\hat{y}_i = y_i],
\]
where $N$ denotes the number of evaluation examples. The primary difference across settings lies in how the score for each option is computed.

\subsubsection{Supervised finetuned models}
For supervised fine-tuning experiments, models were implemented using a
sequence classification head with $|\mathcal{C}|$ output logits,
where $\mathcal{C}$ denotes the set of answer labels (e.g., $\{A, B, C, D\}$).

Given an input example, the model produces logits $z \in \mathbb{R}^{|\mathcal{C}|}$.
The predicted answer is selected as:
\[
\hat{y} = \arg\max_{c \in \mathcal{C}} z_c.
\]

Evaluation accuracy is computed by comparing the predicted label $\hat{y}$ 
with the gold label $y$ over the held-out test set.

\subsubsection{Unsupervised Finetuned models and RAG pipeline}
Rather than generating free-form text, we adopted an MMLU-style label scoring strategy. Given the prompt, we computed the conditional log-probability of each answer option label. At the next-token position, the predicted answer corresponds to the option with the highest log-probability:
\[
\hat{y} = \arg\max_{c \in \mathcal{C}} \log p(c \mid \text{prompt}),
\]
where $\mathcal{C}$ denotes the set of answer labels. $\mathcal{C} = \{A, B, C, D\}$ for 2024 Wikipedia Events dataset and $\mathcal{C} = \{A, B, C, D, E, F, G, H\}$ for QASC dataset.

\section{Results}
For each task and model, we compare four approaches: the base model, retrieval-augmented generation (RAG) using the same base model as the generator, unsupervised fine-tuning (continual pretraining), and supervised fine-tuning.

\subsection{QASC Dataset}
The results on the QASC dataset are reported in Table~\ref{qasc results}. Across all tasks, the base models achieve substantially lower accuracy than both RAG and supervised fine-tuning. Nevertheless, the base models consistently outperform random guessing, which yields an expected accuracy of $\tfrac{1}{C}$ (i.e., $\tfrac{1}{8}$ for QASC). This indicates that the models retain some relevant knowledge from pretraining.

Unsupervised fine-tuning yields only marginal improvements across all three models, suggesting that continual pretraining alone provides limited benefits for multi-hop reasoning. In contrast, RAG improves accuracy by approximately 30 percentage points across all models. Supervised fine-tuning achieves the highest overall accuracy on this dataset.

\subsection{2024 Events Dataset}
The evaluation results on the 2024 Events dataset are summarized in Table~\ref{new events results}. Similar trends are observed across all models. Base models perform poorly on this dataset, reflecting the difficulty of answering questions that rely on temporally novel knowledge. The instruction-tuned model achieves accuracy notably above random guessing, which corresponds to $\tfrac{1}{C}$ (i.e., $\tfrac{1}{4}$), despite lacking explicit access to the underlying event knowledge.

Unsupervised fine-tuning again yields only marginal improvements. In contrast, RAG more than doubles accuracy across all models, highlighting the effectiveness of retrieval-based methods when external, up-to-date knowledge is required. Supervised fine-tuning achieves the largest performance gains overall.

\section{Conclusion}
In this paper, we presented a systematic comparison of parametric and non-parametric knowledge injection mechanisms for open-domain multi-hop question answering. Through controlled experiments on both the QASC benchmark and a newly constructed 2024 Events dataset designed to test temporally novel knowledge, we showed that unsupervised fine-tuning via continual pretraining yields only marginal gains on answer accuracy, suggesting limited effectiveness for improving multi-hop reasoning. In contrast, RAG substantially improves performance, particularly in settings where required knowledge lies beyond the model’s pre-training cutoff, while supervised fine-tuning achieves the highest overall accuracy when task-specific labeled data is available. These findings highlight fundamental differences in how models access and utilize knowledge under different injection strategies and underscore the importance of retrieval-based methods for reasoning over novel or compositional information. Together, our results provide practical guidance for selecting knowledge injection approaches when deploying large language models for multi-hop question answering tasks.

\bibliographystyle{IEEEtran}
\bibliography{refs}

@inproceedings{hotpotqa,
  author    = {Yang, Zhiyuan and Qi, Peng and Zhang, Saizheng and Bengio, Yoshua
               and Cohen, William W. and Salakhutdinov, Ruslan and Manning, Christopher D.},
  title     = {HotpotQA: A Dataset for Diverse, Explainable Multi-hop Question Answering},
  booktitle = {Proceedings of the Conference on Empirical Methods in Natural Language Processing},
  address   = {Brussels, Belgium},
  year      = {2018}
}

@inproceedings{2wikimultihopqa,
  author    = {Ho, Xanh and Nguyen, Anh-Khoa Duong and Sugawara, Saku and Aizawa, Akiko},
  title     = {Constructing a Multi-hop QA Dataset for Comprehensive Evaluation of Reasoning Steps},
  booktitle = {Proceedings of the 28th International Conference on Computational Linguistics},
  address   = {Barcelona, Spain (Online)},
  year      = {2020}
}

@article{musique,
  author  = {Trivedi, Harsh and Balasubramanian, Niranjan and Khot, Tushar and Sabharwal, Ashish},
  title   = {MuSiQue: Multihop Questions via Single-hop Question Composition},
  journal = {Transactions of the Association for Computational Linguistics},
  volume  = {10},
  pages   = {539--554},
  year    = {2022}
}

@inproceedings{rag,
  author    = {Lewis, Patrick and Perez, Ethan and Piktus, Aleksandra and Petroni, Fabio
               and Karpukhin, Vladimir and Goyal, Naman and K{\"u}ttler, Heinrich
               and Lewis, Mike and Yih, Wen-Tau and Rockt{\"a}schel, Tim
               and Riedel, Sebastian and Kiela, Douwe},
  title     = {Retrieval-Augmented Generation for Knowledge-Intensive NLP Tasks},
  booktitle = {Proceedings of the Advances in Neural Information Processing Systems},
  year      = {2020}
}

@inproceedings{dontstopPretraining,
  author    = {Gururangan, Suchin and Marasovi{\'c}, Ana and Swayamdipta, Swabha
               and Lo, Kyle and Beltagy, Iz and Downey, Doug and Smith, Noah A.},
  title     = {Don't Stop Pretraining: Adapt Language Models to Domains and Tasks},
  booktitle = {Proceedings of the Annual Meeting of the Association for Computational Linguistics},
  year      = {2020}
}

@inproceedings{unifiedqa,
  author    = {Khashabi, Daniel and Min, Sewon and Khot, Tushar and Sabharwal, Ashish
               and Tafjord, Oyvind and Clark, Peter and Hajishirzi, Hannaneh},
  title     = {UnifiedQA: Crossing Format Boundaries with a Single QA System},
  booktitle = {Proceedings of the Conference on Empirical Methods in Natural Language Processing},
  year      = {2020}
}

@misc{singleHop_finetuningVSrag,
      title={Fine-Tuning or Retrieval? Comparing Knowledge Injection in LLMs}, 
      author={Oded Ovadia and Menachem Brief and Moshik Mishaeli and Oren Elisha},
      year={2024},
      eprint={2312.05934},
      archivePrefix={arXiv},
      primaryClass={cs.AI},
      url={https://arxiv.org/abs/2312.05934}, 
}

@misc{qascDataset,
      title={QASC: A Dataset for Question Answering via Sentence Composition}, 
      author={Tushar Khot and Peter Clark and Michal Guerquin and Peter Jansen and Ashish Sabharwal},
      year={2020},
      eprint={1910.11473},
      archivePrefix={arXiv},
      primaryClass={cs.CL},
      url={https://arxiv.org/abs/1910.11473}, 
}

@inproceedings{stochastic,
  author    = {Bender, Emily M. and Gebru, Timnit and McMillan-Major, Angelina
               and Shmitchell, Shmargaret},
  title     = {On the Dangers of Stochastic Parrots: Can Language Models Be Too Big?},
  booktitle = {Proceedings of the ACM Conference on Fairness, Accountability, and Transparency},
  year      = {2021}
}

@inproceedings{cot,
  author    = {Wei, Jason and Wang, Xuezhi and Schuurmans, Dale and Bosma, Maarten
               and Ichter, Brian and Xia, Fei and Chi, Ed and Le, Quoc and Zhou, Denny},
  title     = {Chain-of-Thought Prompting Elicits Reasoning in Large Language Models},
  booktitle = {Advances in Neural Information Processing Systems},
  year      = {2022}
}

@inproceedings{Language_Models_as_Knowledge_Bases,
  author    = {Petroni, Fabio and Rockt{\"a}schel, Tim and Riedel, Sebastian
               and Lewis, Patrick and Bakhtin, Anton and Wu, Yuxiang
               and Miller, Alexander},
  title     = {Language Models as Knowledge Bases?},
  booktitle = {Proceedings of the Conference on Empirical Methods in Natural Language Processing},
  year      = {2019}
}

@inproceedings{gpt3,
  author    = {Brown, Tom B. and Mann, Benjamin and Ryder, Nick
               and Subbiah, Melanie and Kaplan, Jared and Dhariwal, Prafulla
               and Neelakantan, Arvind and Shyam, Pranav and Sastry, Girish
               and Askell, Amanda and Agarwal, Sandhini and Herbert-Voss, Ariel
               and Krueger, Gretchen and Henighan, Tom and Child, Rewon
               and Ramesh, Aditya and Ziegler, Daniel M. and Wu, Jeffrey
               and Winter, Clemens and Hesse, Chris and Chen, Mark
               and Sigler, Eric and Litwin, Mateusz and Gray, Scott
               and Chess, Benjamin and Clark, Jack and Berner, Christopher
               and McCandlish, Sam and Radford, Alec and Sutskever, Ilya
               and Amodei, Dario},
  title     = {Language Models are Few-Shot Learners},
  booktitle = {Advances in Neural Information Processing Systems},
  year      = {2020}
}

@misc{flare,
      title={Active Retrieval Augmented Generation}, 
      author={Zhengbao Jiang and Frank F. Xu and Luyu Gao and Zhiqing Sun and Qian Liu and Jane Dwivedi-Yu and Yiming Yang and Jamie Callan and Graham Neubig},
      year={2023},
      eprint={2305.06983},
      archivePrefix={arXiv},
      primaryClass={cs.CL},
      url={https://arxiv.org/abs/2305.06983}, 
}

@misc{Self-rag,
      title={Self-RAG: Learning to Retrieve, Generate, and Critique through Self-Reflection}, 
      author={Akari Asai and Zeqiu Wu and Yizhong Wang and Avirup Sil and Hannaneh Hajishirzi},
      year={2023},
      eprint={2310.11511},
      archivePrefix={arXiv},
      primaryClass={cs.CL},
      url={https://arxiv.org/abs/2310.11511}, 
}

@misc{ChainOFrag,
      title={Chain-of-Retrieval Augmented Generation}, 
      author={Liang Wang and Haonan Chen and Nan Yang and Xiaolong Huang and Zhicheng Dou and Furu Wei},
      year={2025},
      eprint={2501.14342},
      archivePrefix={arXiv},
      primaryClass={cs.IR},
      url={https://arxiv.org/abs/2501.14342}, 
}

@inproceedings{Training_Language_Models_to_Follow_Instructions_with_Human_Feedback,
  author    = {Ouyang, Long and Wu, Jeffrey and Jiang, Xu and Almeida, Diogo
               and Wainwright, Carroll and Mishkin, Pamela and Zhang, Chong
               and Agarwal, Sandhini and Slama, Katarina and Ray, Alex
               and Schulman, John and Hilton, Jacob and Christiano, Paul
               and Leike, Jan and Lowe, Ryan},
  title     = {Training Language Models to Follow Instructions with Human Feedback},
  booktitle = {Advances in Neural Information Processing Systems},
  year      = {2022}
}

@inproceedings{shortcut_learning,
  author    = {Geirhos, Robert and Jacobsen, J{\"o}rn-Henrik and Michaelis, Claudio
               and Zemel, Richard and Brendel, Wieland and Bethge, Matthias
               and Wichmann, Felix A.},
  title     = {Shortcut Learning in Deep Neural Networks},
  booktitle = {Nature Machine Intelligence},
  year      = {2020}
}

@inproceedings{arc,
  author    = {Khashabi, Daniel and Chaturvedi, Snigdha and Roth, Michael
               and Upadhyay, Shyam and Roth, Dan},
  title     = {Question Answering as Global Reasoning over Semantic Abstractions},
  booktitle = {Proceedings of the Conference on Empirical Methods in Natural Language Processing},
  year      = {2018}
}

@misc{openai2024gpt4technicalreport,
      title={GPT-4 Technical Report}, 
      author = {Achiam, John and others},
      year={2024},
      eprint={2303.08774},
      archivePrefix={arXiv},
      primaryClass={cs.CL},
      url={https://arxiv.org/abs/2303.08774}, 
}

@article{deepseek_r1,
  title  = {DeepSeek-R1: Incentivizing Reasoning Capability in Large Language Models},
  author = {DeepSeek-AI and others},
  journal = {arXiv preprint},
  volume = {arXiv:2501.12948},
  year   = {2025}
}

@misc{llama,
      title={LLaMA: Open and Efficient Foundation Language Models}, 
      author={Hugo Touvron and Thibaut Lavril and Gautier Izacard and Xavier Martinet and Marie-Anne Lachaux and Timothée Lacroix and Baptiste Rozière and Naman Goyal and Eric Hambro and Faisal Azhar and Aurelien Rodriguez and Armand Joulin and Edouard Grave and Guillaume Lample},
      year={2023},
      eprint={2302.13971},
      archivePrefix={arXiv},
      primaryClass={cs.CL},
      url={https://arxiv.org/abs/2302.13971}, 
}

@article{llama2_instruction,
  title   = {LLaMA 2: Open Foundation and Fine-Tuned Chat Models},
  author  = {Touvron, Hugo and Martin, Louis and Stone, Kevin
             and Albert, Peter and Almahairi, Amjad and Babaei, Yasmine
             and others},
  journal = {arXiv preprint},
  volume  = {arXiv:2307.09288},
  year    = {2023}
}

@article{mistral,
  title   = {Mistral 7B},
  author  = {Jiang, Albert Q. and Sablayrolles, Alexandre
             and Roux, Antoine and Mensch, Arthur and Savary, Blanche
             and others},
  journal = {arXiv preprint},
  volume  = {arXiv:2310.06825},
  year    = {2023}
}

@misc{Langchain,
      title={Automating Customer Service using LangChain: Building custom open-source GPT Chatbot for organizations}, 
      author={Keivalya Pandya and Mehfuza Holia},
      year={2023},
      eprint={2310.05421},
      archivePrefix={arXiv},
      primaryClass={cs.CL},
      url={https://arxiv.org/abs/2310.05421}, 
}

@misc{BGE,
      title={C-Pack: Packed Resources For General Chinese Embeddings}, 
      author={Shitao Xiao and Zheng Liu and Peitian Zhang and Niklas Muennighoff and Defu Lian and Jian-Yun Nie},
      year={2024},
      eprint={2309.07597},
      archivePrefix={arXiv},
      primaryClass={cs.CL},
      url={https://arxiv.org/abs/2309.07597}, 
}

@article{faiss,
  title={Billion-scale similarity search with {GPUs}},
  author={Johnson, Jeff and Douze, Matthijs and J{\'e}gou, Herv{\'e}},
  journal={IEEE Transactions on Big Data},
  volume={7},
  number={3},
  pages={535--547},
  year={2019},
  publisher={IEEE}
}

@misc{reranker,
      title={When Fine-Tuning Fails: Lessons from MS MARCO Passage Ranking}, 
      author={Manu Pande and Shahil Kumar and Anay Yatin Damle},
      year={2025},
      eprint={2506.18535},
      archivePrefix={arXiv},
      primaryClass={cs.CL},
      url={https://arxiv.org/abs/2506.18535}, 
}

@misc{mmlu,
      title={Measuring Massive Multitask Language Understanding}, 
      author={Dan Hendrycks and Collin Burns and Steven Basart and Andy Zou and Mantas Mazeika and Dawn Song and Jacob Steinhardt},
      year={2021},
      eprint={2009.03300},
      archivePrefix={arXiv},
      primaryClass={cs.CY},
      url={https://arxiv.org/abs/2009.03300}, 
}

\end{document}